\documentclass{article}
\pdfoutput=1

\usepackage{microtype}
\usepackage{graphicx}
\usepackage{subfigure}
\usepackage{booktabs} 
\usepackage{statistics-macros}
\usepackage{grffile}

\newcommand{\riskmax}{\risk_\text{max}}
\newcommand{\alphamin}{\alpha_\text{min}}

\newcommand{\riskdual}{F}
\newcommand{\sB}{\mathcal B}

\newcommand{\riskdro}{\risk_{\rm dro}}

\newcommand{\propvar}{\alpha}
\newcommand{\propk}{\propvar_k}
\newcommand{\propl}{\propvar_l}
\renewcommand{\statrv}{Z}

\usepackage{todonotes}
\usepackage{url}

\usepackage[accepted]{icml2018}

 \newcommand\theTitle{Fairness Without Demographics in Repeated Loss Minimization}
\icmltitlerunning{\theTitle}

\begin{document}

\twocolumn[
\icmltitle{\theTitle}




\begin{icmlauthorlist}
\icmlauthor{Tatsunori B. Hashimoto}{scs,sstat}
\icmlauthor{Megha Srivastava}{scs}
\icmlauthor{Hongseok Namkoong}{sor}
\icmlauthor{Percy Liang}{scs}
\end{icmlauthorlist}

\icmlaffiliation{scs}{Department of Computer Science, Stanford, USA}
\icmlaffiliation{sstat}{Department of Statistics, Stanford, USA}
\icmlaffiliation{sor}{Management Science \& Engineering, Stanford, USA}

\icmlcorrespondingauthor{Tatsunori Hashimoto}{thashim@stanford.edu}

\icmlkeywords{Machine Learning, ICML}

\vskip 0.3in
]



\printAffiliationsAndNotice{}  

\begin{abstract}
Machine learning models (e.g., speech recognizers) are usually trained to minimize average loss,
which results in representation disparity---minority groups (e.g., non-native speakers) contribute less to the training objective and thus tend to suffer higher loss.
Worse, as model accuracy affects user retention,
a minority group can shrink over time.
In this paper, we first show that the status quo of empirical risk minimization (ERM) amplifies representation disparity over time, which can even make initially fair models unfair.
To mitigate this,
we develop an approach based on distributionally robust
optimization (DRO), which minimizes the worst case risk over all distributions
close to the empirical distribution.
We prove that this approach controls the risk of the minority group at each time step,
in the spirit of Rawlsian distributive justice,
while remaining oblivious to the identity of the groups.
We demonstrate that DRO prevents disparity amplification on examples where ERM fails, and
show improvements in minority group user satisfaction in a real-world text autocomplete task.


\end{abstract}

\section{Introduction}

Consider a speech recognizer that is deployed to millions of users. State-of-the art speech recognizers
achieve high overall accuracy, yet
it is well known that such systems have systematically
high errors on minority accents \citep{amodei2016}.
We refer to this phenomenon of high overall accuracy but low minority accuracy as
a \emph{representation disparity}, which is the result of optimizing for average loss. This representation disparity forms our definition of unfairness, and has been observed in face recognition \citep{grother2011},
language identification \citep{blodgett2016, jurgens2017},
dependency parsing \citep{blodgett2016},
part-of-speech tagging \citep{hovy2015},
academic recommender systems \citep{sapiezynski2017},
and automatic video captioning \citep{tatman2017}.

Moreover, a minority user suffering from a higher error rate
will become discouraged and  more likely to stop using the system, thus no longer providing data to the system.
As a result, the minority group will shrink and
might suffer even higher error rates from a retrained model in a future time step.
Machine learning driven feedback loops have been observed in predictive policing \citep{fuster2017predictably} and credit markets \citep{fuster2017predictably}, and this problem of \emph{disparity amplification}
is a possibility in any deployed machine learning system that is retrained on user data. 

In this paper, we aim to mitigate the representation disparity problem and its
amplification through time.  We focus on the following setting: at each time
step, each user interacts with the current model and incurs some loss, based
on which she decides to keep or quit using the service.  A model is trained on
the resulting user data which is used at the next time step.  We assume that
each user comes from one of $K$ groups, and our goal is to minimize the worst
case risk of any group across time.  However, \emph{the group membership and
  number of groups $K$ are both unknown}, as full demographic information is
likely missing in real online services.

We first show that empirical risk minimization (ERM) does not control the
worst-case risk over the disparate $K$ groups and show examples where ERM
turns initially fair models unfair (Section~\ref{sec:erm}). To remedy this
issue, we propose the use of distributionally robust optimization (DRO)
(Section~\ref{sec:dro}). Given a lower bound on the smallest group
proportion, we show that optimizing the worst-case risk over an appropriate
chi-square divergence ball bounds the worst-case risk over
groups. Our approach is computationally efficient, and can be applied as a
small modification to a wide class machine learning models
trained by stochastic gradient descent methods. We show that DRO succeeds on
the examples where ERM becomes unfair, and demonstrate higher average
minority user satisfaction and lower disparity amplification on a Amazon
Mechanical Turk based autocomplete task.


\subsection{Fairness in Machine Learning}

Recently, there has been a surge of interest in fairness in machine learning
\cite{barocas2016}.  Our work can be seen as a direct instantiation of John
Rawls' theory on distributive justice and stability, where we view predictive
accuracy as a resource to be allocated.  Rawls argues that the
\emph{difference principle}, defined as maximizing the welfare of the
worst-off group, is fair and stable over time since it ensures that minorities
consent to and attempt to maintain the status quo \citep[p155]{rawls2001}.

In this work, we assume the task is general loss minimization, and demographic data is unavailable.
This differs from the substantial
body of existing research into fairness for classification problems involving
protected labels such as the use of race in recidivism protection
\cite{chouldechova2017}.
There has been extensive work \cite{barocas2016} on guaranteeing fairness for
classification over a protected label through constraints such as equalized
odds \cite{woodworth2017, hardt2016}, disparate impact \cite{feldman2015} and
calibration \cite{kleinberg2017}.  However, these approaches require the use
of demographic labels, and are designed for classification tasks.  This makes
it difficult to apply such approaches to mitigate representation disparity in
tasks such as speech recognition or natural language generation where full
demographic information is often unavailable.

A number of authors have also studied individual notions of fairness, either
through a fixed similarity function \cite{dwork2012} or subgroups of a set of
protected labels \cite{kearns2018gerrymandering, hebertjohnson2017}.
\citet{dwork2012} provides fairness guarantees without explicit groups, but
requires a fixed distance function which is difficult to define for real-world
tasks. \citet{kearns2018gerrymandering, hebertjohnson2017} consider subgroups
of a set of protected features, but defining non-trivial protected features
which cover the latent demographics in our setting is difficult. Although
these works generalize the demographic group structure, similarity and
subgroup structure are both ill-defined for many real-world tasks.

In the online setting, works on fairness in bandit learning \cite{joseph2016,
  jabbari2017} propose algorithms compatible with Rawls' principle on equality
of opportunity---an action is preferred over another only if the true quality
of the action is better. Our work differs in considering Rawlsian fairness for
distributive justice \cite{rawls2009}.  Simultaneous with our work,
\citet{liu2018delayed} analyzed fairness over time in the context of
constraint based fairness criteria, and show that enforcing static fairness
constraints do not ensure fairness over time. In this paper, we consider
latent demographic groups and study a loss-based approach to fairness and
stability.

\vspace{-3pt}
\section{Problem setup}
\label{sec:problem}

We begin by outlining the two parts of our motivation: \emph{representation disparity}
and \emph{disparity amplification}.

\textbf{Representation disparity}: Consider the standard loss-minimization
setting where a user makes a query $\statrv \sim P$, a model
$\param \in \Theta$ makes a prediction, and the user incurs loss
$\loss(\param; \statrv)$. We denote the expected loss as the risk
$\risk(\param) = \E_{Z \sim P}[\loss(\theta; Z)]$. The observations $\statrv$
are assumed to arise from one of $K$ latent groups such that
$\statrv \sim P \defeq \sum_{k\in [K]} \propk P_k$.  We assume that neither
the population proportions $\{\propk\}$ nor the group distributions $\{P_k\}$
are known.
The goal is to control the worst-case risk over all $K$ groups:
\begin{equation}\label{eq:staticrisk}
  \riskmax(\param) = \max_{k \in [K]} \risk_k(\param), \quad \risk_k(\param) \defeq \E_{P_k}[\loss(\param; \statrv)].
\end{equation}
Representation disparity refers to the phenomenon of low $\risk(\param)$ and high $\riskmax(\param)$ due to a group with small $\alpha_k$.

\textbf{Disparity amplification:} To understand the amplification of
representation disparity over time, we will make several assumptions on the
behavior of users in response to observed losses. These assumptions are
primarily for clarity of exposition---we will indicate whenever the
assumptions can be relaxed leave generalizations to the supplement. Roughly
speaking, minimizing the worst-case risk $\riskmax(\theta)$ should mitigate
disparity amplification as long as lower losses lead to higher user
retention. We now give assumptions that make this intuition precise.

In the sequential setting, loss minimization proceeds over $t=1, 2, \hdots T$
rounds, where the group proportion $\propk^{(t)}$ depends on $t$ and varies
according to past losses. At each round $\lambda_k^{(t+1)}$ is the expected
number of users from group $k$, which is determined by $\nu(\risk_k(\theta))$,
the fraction of users retained, and $b_k$, the number of new users (see
Definition \ref{def:dynamics}). Here, $\nu$ is a differentiable, strictly
decreasing retention function which maps a risk level $\risk$ to the fraction
of users who continue to use the system. Modeling user retention as a
decreasing function of the risk implies that each user makes an independent
decision of whether to interact with the system at time $t+1$ based on their
expected loss at time $t$. For example, selecting $\nu(x) = 1-x$ and $\risk_k$
equal to the expected zero-one loss implies that users leave proportional to
the misclassification rates of their queries.

At each round we learn parameters $\param^{(t+1)}$ based on
$n^{(t+1)}\sim \text{Pois}(\sum_k \lambda_k^{(t+1)})$ users (data points).
While we define the sample size as a Poisson process for concreteness,
our main results hold for any distribution fulfilling the strong law of large
numbers, as we perform all stability analyses in the population limit. 
\begin{definition}[Dynamics]
  \label{def:dynamics}
  Given a sequence $\theta^{(t)}$, for each $t = 1\hdots T$,
  the expected number of users $\lambda$ and samples $Z_i^{(t)}$ starting at $\lambda_k^{(0)}=b_k$ is governed by:
  \begin{align*}
  \lambda_k^{(t+1)} &\defeq  \lambda_k^{(t)} \nu(\risk_k(\param^{(t)}))+b_k\\
    \alpha_k^{(t+1)} &\defeq
                       \frac{\lambda_k^{(t+1)}}{\sum_{k'\in[K]} \lambda_{k'}^{(t+1)}}
  \end{align*}
  \begin{align*}
    n^{(t+1)} &\defeq \text{Pois}(\sum_k \lambda_k^{(t+1)}) \\
    \statrv_1^{(t+1)} \hdots \statrv_{n^{(t+1)}}^{(t+1)}
                   & \simiid P^{(t+1)} \defeq \sum_{k\in[K]} \alpha_k^{(t+1)} P_k.
\end{align*}
\end{definition}
If we use ERM at each time step the parameter sequence is defined as
$\param^{(t)} = \arg \min_{\param \in \Theta} \sum_i \loss(\param;
\statrv_i^{(t)})$.


Our goal is to control over all groups $k = 1, \ldots, K$ and time periods $t = 1, \ldots, T$ the group-wise risk $\risk_k(\theta^{(t)})$,
\begin{equation}
  \label{eq:obj}
  \riskmax^T(\theta^{(0)}, \cdots, \theta^{(T)})
  = \max_{k,t} \left\{\risk_k(\param^{(t)})\right\}.
\end{equation}

Without knowledge of group membership labels, population proportions
$\propk^{(t)}$, new user rate $b_k$, and retention rate $\nu$, minimizing
$\riskmax^T$ gives rise to two major challenges. First, without group
membership labels there is no way to directly measure the worst-case risk
$\riskmax^T$, let alone minimize it. Second, we must ensure that the group
proportions $\alpha_k^{(t)}$ are stable, since if $\alpha_k^{(t)} \to 0$ as
$t \to \infty$ for some group $k \in [K]$, then no algorithm can control
$\riskmax^T$ when a group has near zero probability of appearing in our
samples.

We begin by illustrating how models that are initially fair with low
representation disparity may become unfair over time if we use ERM
(Section~\ref{sec:erm}). We then propose a solution based on distributionally
robust optimization (Section~\ref{sec:dro}), and study examples where this approach
mitigates representation disparity in our experimental section (Section~\ref{sec:experiments}).


\section{Disparity amplification}
\label{sec:erm}

The standard approach to fitting a sequence of models $\param^{(t)}$ is to
minimize an empirical approximation to the population risk at each time
period. In this section, we show that even minimizing the population risk
fails to control minority risk over time, since expected loss (average case)
leads to disparity amplification. The decrease in user retention for the
minority group is exacerbated over time since once a group shrinks sufficiently,
it receives higher losses relative to others, leading to even fewer
samples from the group.

\subsection{Motivating example}
\begin{figure}[ht]
        \vspace{-5pt}
  \centering
  \includegraphics[scale=0.35]{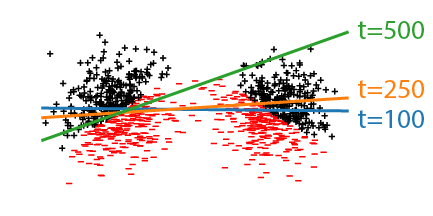}
      \vspace{-8pt}
  \caption{An example online classification problem which begins fair, but
    becomes unfair over time.}
  
  \label{fig:exone}
\end{figure}
 
Consider the two-class classification problem in Figure~\ref{fig:exone} where
the two groups are drawn from Gaussians and the optimal classification
boundary is given along $x_2 = 0$. Assume that the sampling distribution evolves
according to definition \ref{def:dynamics} with $\nu(x) = 1.0-x$, $\loss$
equal to the zero one loss, and $b_0=b_1=n^{(0)}_0=n^{(0)}_1=1000$. Initially,
ERM has similar and high accuracy on both groups with the boundary $x_2>0$,
but over time random fluctuations in accuracy result in slightly fewer samples
from the cluster on the right. This leads to disparity amplification since ERM
will further improve the loss on the \emph{left} cluster at the expense of the
\emph{right} cluster.  After 500 rounds, there are nearly no samples from the
\emph{right} cluster, and as a result, the \emph{right} cluster ends up suffering
high loss.


\subsection{Conditions for disparity amplification}
\label{sec:fairstab}

The example above demonstrated that disparity amplification can occur easily
even in a situation where the two groups have identical population size and
initial risk. In general if we view the expected
user counts $\lambda^{(t)}$ as a dynamical system, the long-term fairness
properties for any fairness criteria are controlled by two factors -
whether $\lambda$ has a fair fixed
point (defined as a population fraction where risk minimization maintains the
same population fraction over time) and whether this fixed point
is stable.

Fixed points of risk minimization are determined by a
combination of user retention function $\nu$ and the models $\theta^{(t)}$,
and without knowledge of $\nu$ it is hard to ensure that a model has a fair
fixed point. Even if a fixed point is fair, such as when the population fraction
and risk received by each group is equal, and we start at this fair fixed
point, minimizing the empirical loss may deviate from this fair fixed
point over time due to finite sample fluctuations or noise in the model estimation procedure. 

To show this result, we study the dynamical system $\Phi$, which is defined by
dynamics in Definition \ref{def:dynamics} with $\theta$ derived from
minimizing the population, rather than empirical risk. 
\begin{definition}
  Let $\Phi$ be the update for the expected population size
  \[\lambda^{(t+1)}_k \defeq \Phi(\lambda^{(t)}_k) = \lambda^{(t)}_k \nu(\risk_k(\param(\lambda^{(t)}_k)))+ b_k,\]
  \[\param(\lambda^{(t)}_k) = \arg\min_\param \E_{\sum_k \propk^{(t)}P_k} [ \loss(\param; \statrv)].\]
\end{definition}

The arrival intensity $\lambda^*$ is called a fixed point if
$\lambda^* = \Phi(\lambda^*)$. This fixed point is \emph{stable} whenever the
maximum modulus of the eigenvalues of the Jacobian of $\Phi$ is less than one and \emph{unstable} whenever it is greater than one \citep[Theorem 2.1]{luo2012regularity}. 

Proposition~\ref{prop:unstable} gives a precise statement of this
phenomenon. We prove the result in Section~\ref{sec:proof-of-instability}, and
further show a generalization to general dynamics
$\Phi(\lambda_k) = h(\lambda_k, \risk_k)$ where $h$ is differentiable and
monotone in the second argument. We denote by $\rho_{\rm max}(A)$ the maximum
modulus of the eigenvalues of $A$.
\begin{proposition}
  \label{prop:unstable}
  Let $\lambda^* = \Phi(\lambda^*)$ be a fixed point, and
  $\theta^* = \arg\min_\theta \E_{\sum_k \propk^* P_k}[\loss(\param;
  \statrv)]$ be the minimizer at $\lambda^*$.
  
  Define $H_{\risk}(\propvar^*)$ as the positive definite Hessian of the
  expected risk at $\theta^*, \lambda^*$ and define $\nabla L$ as the per-group parameter gradients at
  $\theta^*$,
  \[\nabla L = \begin{bmatrix}
      \nabla_\param \E_{P_1}[\loss(\param^*; \statrv)] \\
      \vdots\\
      \nabla_\param \E_{P_k}[\loss(\param^*; \statrv)]\\
    \end{bmatrix}.
  \]
  The arrival intensity $\lambda^*$ is unstable whenever
  \begin{multline*}
    \rho_{\rm max}\bigg( \diag(\nu(\risk(\theta(\lambda^*)))) 
    - \diag(\lambda^* \nu'(\risk(\theta(\lambda^*)) \\
    \nabla L H_{\risk}(\propvar^*)^{-1} \nabla L^\top \left( \frac{I}{\sum_k \lambda^*_k} - \frac{\mathbf{1} \lambda^{*\top}}{(\sum_k\lambda^*_k)^2}\right) \bigg) > 1.
  \end{multline*}
\end{proposition}
We see that the major quantities which control risk are the retention rate $\nu$ and its derivative, as well as a $K \times K$ square matrix $\nabla L H_{\risk}(\propvar^*)^{-1} \nabla L^\top$ which roughly encodes the changes in one group's risk as a function of another.

We can specialize the stability condition to obtain an intuitive and negative
result for the stability of risk minimization (average case). Even if we start
at a fair fixed point with $\lambda^*_1 = \cdots =\lambda^*_k$ and
$\risk_1 = \cdots =\risk_k$,  if decreasing the risk for one group
increases the risk for others sufficiently, the fixed point is unstable and
the model will eventually converge to a different, possibly unfair, fixed
point.

\begin{corollary}[Counterexample under symmetry]\label{cor:symmetry}

  Let $\lambda^*_1 = \cdots = \lambda^*_k$ be a fixed point with
  $\risk_1 = \cdots = \risk_k$, then for any strongly convex loss,
  \begin{equation}
    \label{eq:instability}
    \rho_{\rm max}\bigg( \nabla L H_{\risk}(\propvar^*)^{-1} \nabla L^\top 
    \bigg) > \frac{1 - \nu(\risk_1)}{- \nu'(\risk_1)/k}.
  \end{equation}
  is a sufficient condition for instability.
\end{corollary}
See Section~\ref{sec:proof-of-symmetry} for proof and generalizations.

The bound~\eqref{eq:instability} has a straightforward interpretation. The
left hand side is the stability of the model, where maximal eigenvalue of the matrix
$\nabla L H_{\risk}(\propvar^*)^{-1} \nabla L^\top$ represents the maximum excess risk
that can be incurred due to a small perturbation in the mixture weights $\alpha$.
The right hand side represents the underlying stability of the dynamics and
measures the sensitivity of $\lambda$ with respect to risk.

\textbf{Mean and median estimation:} Consider a simple mean estimation example
where each user belongs to one of two groups, $-1$ or $1$ and incurs loss
$(\param - \statrv)^2$. $\param=0$ is clearly a fair fixed point, since it
equalizes losses to both groups, with $H_{risk}(\propvar^*) = 1/2$ and
$\nabla L = [2, -2]$ making
$\rho_{\rm max}\bigg( \nabla L H_{\risk}(\propvar^*)^{-1} \nabla L^\top \bigg)
= 4$. If we select $\nu(x) = \exp(-x)$, the right hand side becomes
$2(1-e^{-1})e \approx 3.4$, and thus any perturbation will eventually result
in $\lambda_1 \neq \lambda_2$. In this case the only other fixed points are
the unfair solutions of returning the mean of either one of the groups.

The situation is even worse for models which are not strongly convex, such as
median estimation. Replacing the squared loss above with the absolute value
results in a loss which has a non-unique minimizer at $0$ when
$\lambda_1 = \lambda_2$ but immediately becomes $-1$ whenever
$\lambda_1 > \lambda_2$. In this case, no conditions on the retention function
$\nu$ can induce stability. This fundamental degeneracy motivates us to search
for loss minimization schemes with better stability properties than ERM
(average case).

\section{Distributionally robust optimization (DRO)}
\label{sec:dro}

Recall that our goal is to control the worst-case risk~\eqref{eq:obj} over all
groups and over all time steps $t$.  We will proceed in two steps.
First, we show that performing distributionally robust optimization controls
the worst-case risk $\riskmax(\param^{(t)})$ for a single time step.
Then, we show that this results in a lower bound on group proportions
$\{ \alpha_k^{(t)} \}_{k=1}^K$, and thus ensures control over the worst-case
risk for all time steps.
As a result of the two steps, we show in Section~\ref{sec:stability} that our
procedure mitigates disparity amplification over \emph{all time steps}. For
notational clarity, we omit the superscript $t$ in
Sections~\ref{sec:bound-risk}-\ref{sec:optimization}.

\subsection{Bounding the risk over unknown groups}
\label{sec:bound-risk}

The fundamental difficulty in controlling the worst-case group risk over a
single time-step~$\riskmax(\theta^{(t)})$ comes from not observing the group
memberships from which the data was sampled. For many machine learning systems
such as speech recognition or machine translation, such situations are common
since we either do not ask for sensitive demographic information, or it is
unclear a priori which demographics should be protected. To achieve reasonable
performance across different groups, we postulate a formulation that protects
against \emph{all} directions around the data generating distribution. We
build on the distributionally robust formulation of~\citet{DuchiGlNa16} which
will allow us to control the worst-case group risk $\riskmax(\theta^{(t)})$.

To formally describe our
approach, let $\dchi{P}{Q}$ be the $\chi^2$-divergence between probability
distributions $P$ and $Q$ given by
$\dchi{P}{Q} \defeq \int \left( \frac{dP}{dQ} - 1 \right)^2 dQ$. If $P$ is not
absolutely continuous with respect to $Q$, we define $\dchi{P}{Q} \defeq \infty$.

Let $\sB(P, r)$ be the chi-squared ball around a probability distribution
$P$ of radius $r$ so that $\sB(P, r) \defeq \{ Q \ll P : \dchi{Q}{P} \le r
\}$. We consider the worst-case loss over all
$r$-perturbations around $P$,
\begin{equation}
  \label{eq:dro}
  \riskdro (\param; r) \defeq \sup_{Q \in \sB(P, r)} \E_Q[\loss(\theta; Z)].
\end{equation}
Intuitively, the distributionally robust risk $\riskdro(\theta;
r)$ upweights examples $Z$ with high loss $\loss(\theta;
Z)$. If there is a group suffering high loss, the corresponding mixture
component will be over-represented (relative to the original mixture weights)
in the distributionally robust risk $\riskdro(\theta;
r)$.
We show in the following proposition that $\riskdro(\theta;
r)$ bounds the risk of each group
$\risk_k(\theta)$, and hence the group-wise worst-case
risk~\eqref{eq:staticrisk}, for an appropriate choice of the robustness
radius $r$.


\begin{proposition}
  \label{prop:dro-includes-mixture}
  For $P \defeq \sum_{k\in [K]} \propk P_k$, we have
  $\risk_k(\param) \le \riskdro(\param; r_k)$ for all $\theta \in \Theta$
  where $r_k := \left(1/\alpha_k-1\right)^2$ is the robustness radius.
\end{proposition}
We prove the result in
Section~\ref{sec:proof-of-dro-includes-mixture}. Roughly speaking, the above
bound becomes tighter if the variation in the loss $\loss(\theta; Z)$ is
substantially higher between groups than within each group. In particular,
this would be the case if the loss distribution for each group have distinct
support with relatively well-concentrated components within each group.

As a consequence of Proposition~\ref{prop:dro-includes-mixture}, if we have a
lower bound on the group proportions
$\alphamin \le \min_{k \in [K]} \alpha_k$, then we can control the worst-case
group risk $\riskmax(\theta)$ by minimizing the upper bound
$\theta \mapsto \riskdro(\theta; r_{\rm max})$ where
$r_{\rm max} \defeq (1/\alphamin - 1)^2$.

Similar formulations for robustness around the empirical distribution with
radius shrinking as $r / n$ had been considered in ~\citep{Ben-TalHeWaMeRe13,
  LamZh15, DuchiNa16}. While there are many possible robustness balls $\sB$
which could provide upper bounds on group risk, we opt to use the chi-squared
ball since it is straightforward to optimize \citep{Ben-TalHeWaMeRe13,
  NamkoongDu16, NamkoongDu17} and we found it empirically outperformed other
$f$-divergence balls.






  
\subsection{Interpreting the dual}
\label{sec:dual}
The dual of the maximization problem~\eqref{eq:dro} provides additional
intuition on the behavior of the robust risk. 




\begin{proposition}[\citep{DuchiNa18}]
  \label{prop:dual}
  If $\loss(\theta; \cdot)$ is upper semi-continuous for any $\theta$, then
  for $r_{\rm max} \ge 0$ and any $\theta$, $\riskdro(\theta; r_{\rm max})$ is
  equal to the following expression
  \begin{align}
    \label{eq:dual}
    \inf_{\eta \in \R} \left\{
    \riskdual(\theta; \eta) \defeq
    C \left(\E_P\left[\hinge{\loss(\theta, \statrv) - \eta}^2\right]\right)^{\half} + \eta
    \right\}
  \end{align}
  where
  $C = \left(2 (1/\alphamin - 1)^2 + 1\right)^{1/2}$.
\end{proposition}
Denoting by $\eta\opt$ the optimal dual variable~\eqref{eq:dual}, we see from
the proposition that all examples suffering less than $\eta\opt$-levels of
loss are completely ignored, and large losses above $\eta\opt$ are upweighted
due to the squared
term. 

However, unlike standard parameter regularization techniques, which
encourage $\param$ to be close to some point, our objective biases the model to have
 fewer high loss examples which matches our goal of mitigating representation
disparity.

\begin{figure}[h!]
      \vspace{-5pt}
  \centering
  \includegraphics[scale=0.33]{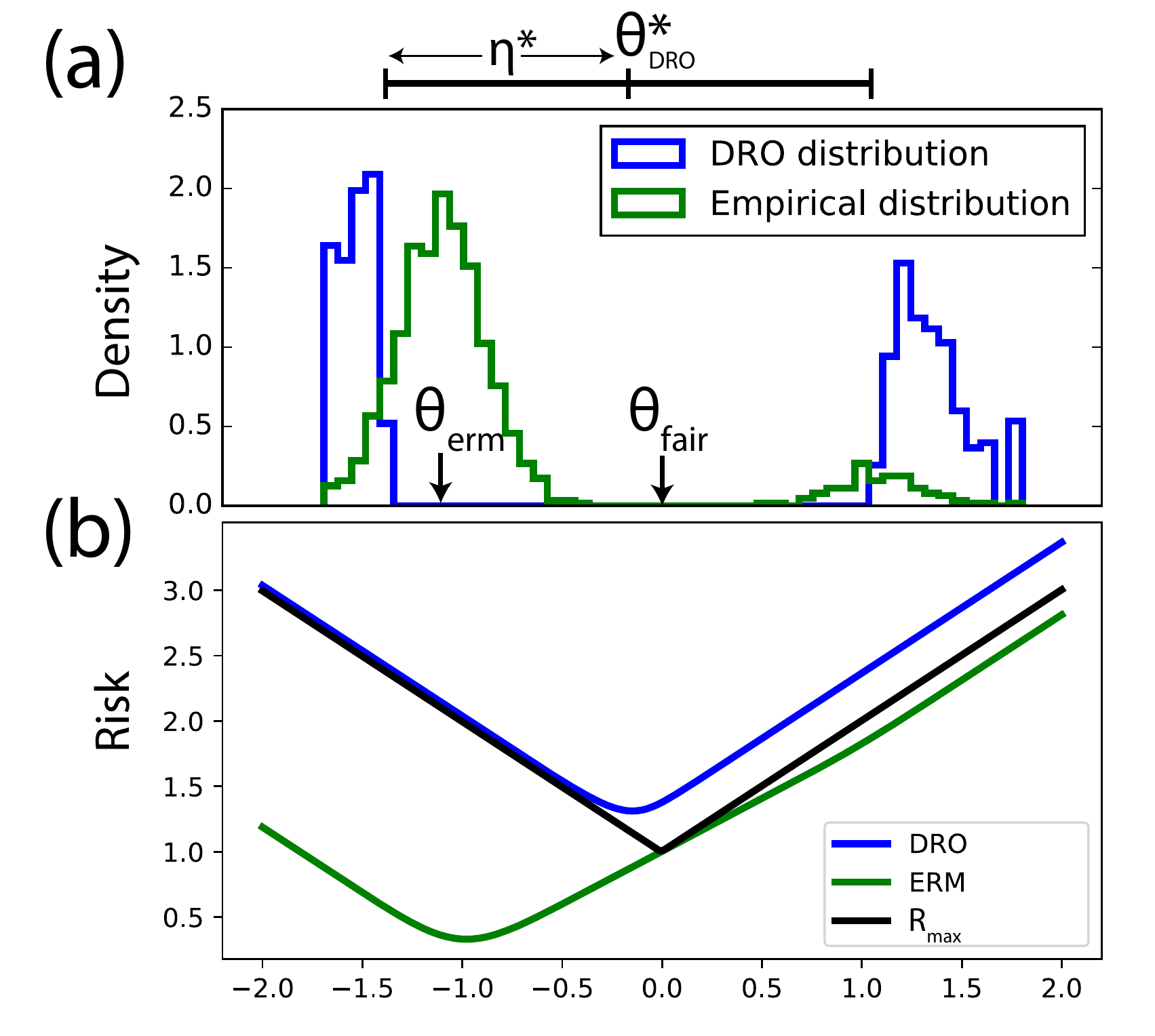}
    \vspace{-8pt}
  \caption{Chi-square distributionally robust optimization (DRO) regularizes the losses (top panel) such that the minimum loss estimate is fair to both groups (bottom panel).
  }

  \label{fig:medest}
\end{figure}

\textbf{Median Estimation}: Recall the median estimation problem over two
groups mentioned in Section \ref{sec:fairstab} where the loss is
$\loss(\param; \statrv) = \|\param-\statrv\|_1$.  Figure \ref{fig:medest}
shows the behavior of both ERM and DRO on this median estimation task with
unbalanced ($\alpha_{\min} = 0.1$) groups.  The parameter estimate which
minimizes $\riskmax$ for this problem is $\theta_{\text{fair}}=0$ since this
is equidistant from both groups. ERM on the other hand focuses entirely on the
majority and returns $\theta_{\text{ERM}}\approx -1.0$.

DRO returns $\theta^*_{\text{DRO}}$ which is close to
$\theta_{\text{fair}}$. Analyzing the risk, we find that the single-step worst-case group risk $\riskmax(\theta)$
 in~\eqref{eq:staticrisk} is an
upper bound on ERM, and DRO forms a tight upper bound this quantity (Figure
\ref{fig:medest}b).  We can also understand the behavior of DRO through the
worst-case distribution $Q$ in Equation \ref{eq:dro}. Figure \ref{fig:medest}a
shows the worst-case distribution $Q$ at the minimizer $\theta^*_{\text{DRO}}$
which completely removes points within distance $\eta^*$. Additionally, points
far from $\theta^*_{\text{DRO}}$ are upweighted, resulting in a large
contribution to the loss from the minority group.

We expect the bound to be tight when all individuals within a group receive the same loss. In this case, thresholding
by $\eta^*$ corresponds to selecting the single highest risk group
which is equivalent to directly minimizing $\riskmax(\theta)$~\eqref{eq:staticrisk}.

On the other hand, the worst case for our approach is if $\alpha_{\min}$ is small, and a group with low expected loss has a high loss tail with population size $\alpha_{\min}$. In this case DRO is a loose upper bound and optimizes the losses of the group with already low expected loss.

This is closely related to recent observations that the DRO bound can be loose for classification losses such as the zero-one loss due to the worst-case distribution consisting purely of misclassified examples \cite{hu2018does}. Even in this case, the estimated loss is still a valid upper bound on the worst case group risk, and as Figure \ref{fig:medest} shows, there are examples where the DRO estimate is nearly tight.

\subsection{Optimization}
\label{sec:optimization}

We now show how to minimize 
$\theta \mapsto \riskdro(\theta; r_{\rm max})$ efficiently for a large class of problems. For models such as deep neural networks that rely on stochastic gradient descent, the dual objective $\riskdual(\param; \eta)$ in~\eqref{eq:dual} can be used directly since it only involves an expectation over
the data generating distribution $P$.

Formally, the following  procedure optimizes \eqref{eq:dro}: for a given value of
$\eta$, compute the approximate minimizer $\what{\theta}_\eta$
\begin{equation}
  \label{eq:sgd}
  \minimize_{\param \in \Theta} \E_{P}\hinge{\loss(\param;\statrv)-\eta}^2.
\end{equation}
From Propositions~\ref{prop:dro-includes-mixture} and~\ref{prop:dual}, we have
\begin{equation*}
  \riskmax(\what{\theta}_\eta)
  \le \riskdro(\what{\theta}_\eta; r_{\rm max})
  \le \riskdual(\what{\theta}_{\eta}, \eta)
\end{equation*}
which implies that we can treat $\eta$ as a hyperparameter.
For convex losses $\theta \mapsto \loss(\theta; Z)$,
the function
$\eta \mapsto \riskdual(\what{\theta}_{\eta}, \eta)$ is convex,
and thus we can perform a binary search over $\eta$ to find the global optimum efficiently.


Alternatively, for models where we can compute
$\theta^*(Q) \in \argmin_{\theta \in \Theta} \E_Q[\loss(\param; Z)]$ efficiently, we can
use existing primal solvers that compute the worst-case probability distribution
$Q^*(\theta) \in \argmax_{Q \in \sB(P, r)} \E_Q[\loss(\theta; Z)]$ for a given
$\theta$ based on projected gradient ascent on $Q$ ~\citep{NamkoongDu16}. By
alternating between optimization on $\theta$ and $Q$, we can efficiently find
the saddle point $(\theta^*, Q^*)$ that satisfies $\theta^* = \theta^*(Q^*)$
and $Q^* = Q^*(\theta^*)$.






\subsection{Stability of minority loss minimization}
\label{sec:stability}

We have thus far demonstrated that for a single time step, the worst-case risk
over all groups $\riskmax(\theta) = \max_k \risk_k(\theta)$ can be controlled
by the distributionally robust risk $\riskdro(\theta; r_{\rm max})$ where
$r_{\rm max} \defeq (1/\alphamin - 1)^2$ and $\alphamin$ is the minority group
proportion. Now, we study how the individual group risk $\risk_k(\theta)$
affects user retention and hence \emph{future} risk. By virtue of
providing an upper bound to $\riskmax(\theta)$, optimizing
$\riskdro(\theta; r_{\rm max})$ at each time step can thus control the
\emph{future} group risk $\riskmax(\theta)$.

We show that if the initial group proportions satisfy
$\propk^{(0)} \ge \alphamin$ and the worst-case risk $\riskmax(\theta^{(t)})$
is sufficiently small at each time $t$, then we can ensure
$\propk^{(t+1)} > \alphamin$. Thus, to control $\riskmax^{T}$, the worst-case
group risk over \emph{all time steps}, it suffices to control
$\riskdro(\theta^{(t)}; r_{\rm max})$ using the procedure in
Section~\ref{sec:optimization}.
\begin{proposition}
  \label{prop:dynamics}
  Assume the retention model in Definition~\ref{def:dynamics}. Let
  $\propk^{(t)} > \alphamin$, $\frac{b_k}{\sum_k b_k} > \alphamin$, $\lambda^{(t)}\defeq \sum_k \lambda_k^{(t)} \leq \frac{\sum_k b_k}{1-\nu_{\max}}$, and
  $\nu(\risk_k(\theta^{(t)})) < \nu_{\max}$. Then, whenever we have
  \[ \risk_k(\param^{(t)}) \leq \nu^{-1}\left(1 -
      \frac{(1-\nu_{max})b_k}{\alphamin \sum_k b_k}\right),\]
  \[\propk^{(t+1)}  = \frac{\lambda^{(t)}\propk^{(t)} \nu(\risk_k(\param^{(t)}))+b_k}{\sum_l \lambda^{(t)}\propl^{(t)} \nu(\risk_l(\param^{(t)}))+b_l} > \alphamin.\]
\end{proposition}

We conclude that as long as we can guarantee
\begin{equation}
  \label{eq:dro-small}
  \riskdro(\theta^{(t)}; r_{\rm max}) \le \nu^{-1}\left(1 -
    \frac{(1-\nu_{max})b_k}{\alphamin \sum_k b_k}\right),
\end{equation}
we can control $\riskmax^T(\theta^{(0)}, \ldots, \theta^{(T)})$, the unknown
worst-case group risk over \emph{all time steps} by optimizing
$\riskdro(\theta^{(t)}; r_{\rm max})$ at each step $t$. While the
condition~\eqref{eq:dro-small} is hard to verify in practice, we observe
empirically in Section~\ref{sec:experiments} that optimizing the
distributionally robust risk $\riskdro(\theta^{(t)}; r_{\rm max})$ at time
step $t$ indeed significantly reduces disparity amplification in comparison to
using ERM.

Proposition~\ref{prop:dynamics} gives stronger fairness guarantees than the
stability conditions for ERM in Proposition~\ref{prop:unstable}. In ERM the
best one can do is to add strong convexity to the model to stabilize to a
possibly unfair fixed point.  In contrast, Proposition~\ref{prop:dynamics}
gives conditions for controlling $\riskmax$ over time without
assuming that there exists a fair fixed point.


\textbf{Stability of median estimation:} Returning to our running example of geometric median
estimation, we can show that under the same dynamics, ERM is highly unstable while
DRO is stable. Consider a three Gaussian mixture on the corners of the simplex, with
$L_2$ loss, retention function $\nu(r) = \exp(-r)$, and $b_1=b_2=50$, $n^{(t)}=1000$. By construction,
$(1/3,1/3,1/3)$ is the fair parameter estimate.

\begin{figure}[h!]
  \centering
  \subfigure[ERM]{\includegraphics[trim={1.5cm 0 0 1.5cm},scale=0.16]{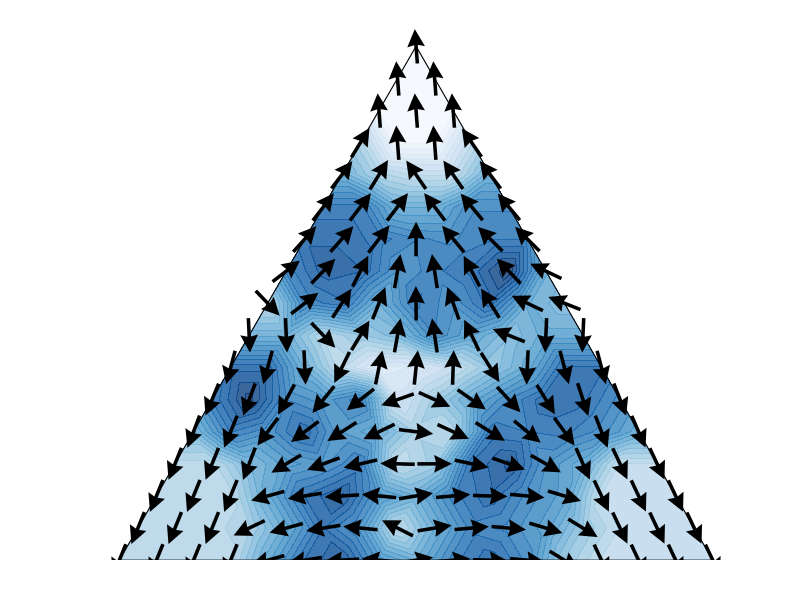}}
  \phantom{--}
  \subfigure[DRO ($\propk=0.4$)]{\includegraphics[trim={1.5cm 0 0 1.5cm}, scale=0.16]{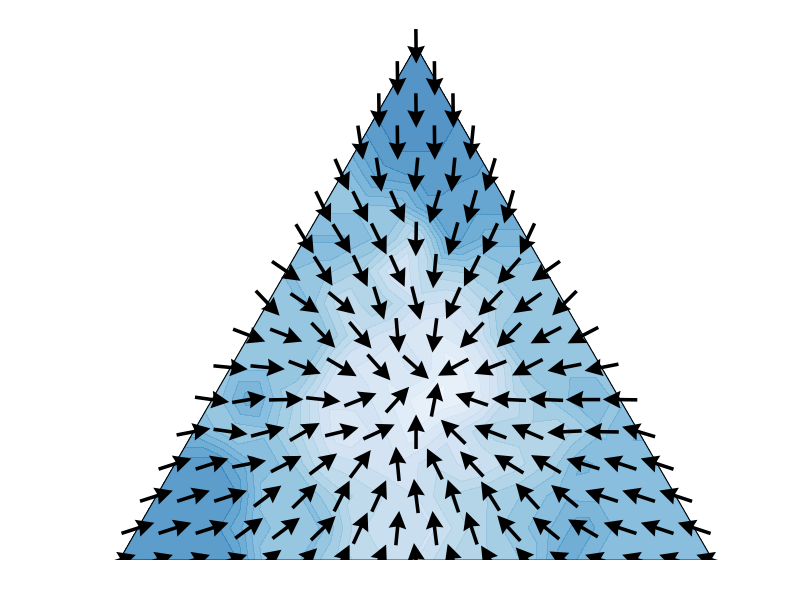}}
  \vspace{-5pt}
  \caption{Dynamics of repeated median estimation - shading indicates velocity at each point. ERM  results
    in unfair parameter estimates that favor one group. DRO is strongly stable, with an equal proportion groups being the unique stable equilibrium. }
    \vspace{-5pt}
  \label{fig:meddyn}
  \end{figure}

Figure \ref{fig:meddyn} shows that ERM is highly unstable,
with the only stable fixed points being the corners, where a single group dominates all others.
The fair parameter estimate is an unstable fixed point for ERM, and any perturbation eventually
results in a completely unfair parameter estimate.
On the other hand, DRO has the reverse behavior, with the fair parameter estimate
being the unique stable fixed point.



\begin{figure*}[h]
  \vspace{-5pt}
  \centering
  \subfigure[User satisfaction]{\includegraphics[scale=0.5]{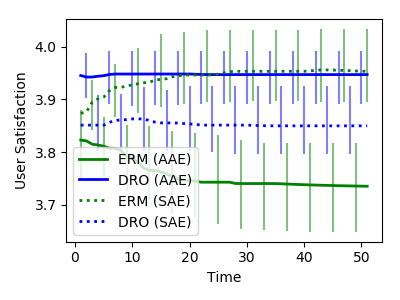}}
  \subfigure[User retention]{\includegraphics[scale=0.5]{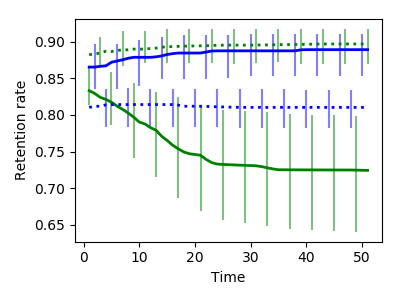}}
  \subfigure[User count]{\includegraphics[scale=0.5]{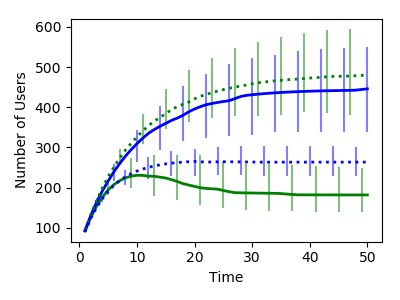}}
  \vspace{-5pt}
  \caption{Inferred dynamics from a Mechanical Turk based evaluation of autocomplete systems. DRO increases minority (a) user satisfaction and (b) retention, leading to a corresponding increase in (c) user count. Error bars indicates bootstrap quartiles.}
\vspace{-5pt}
\label{fig:auto1}

\end{figure*}

\section{Experiments}
\label{sec:experiments}

We demonstrate the effectiveness of DRO on our motivating example (Figure
\ref{fig:exone}) and human evaluation of a text autocomplete system on Amazon
Mechanical Turk. In both cases, DRO controls the worst-case risk $\riskmax^T$
over time steps and improves minority retention.

\subsection{Simulated task}

Recall  the motivating example in Figure \ref{fig:exone} which shows that logistic regression applied
to a two-class classification problem is unstable and becomes pathologically unfair.

The data is constructed by drawing from a mixture of two Gaussians (groups)
centered at $(-1.5,0)$ and $(0,1.5)$.
The two groups are labeled according to the linear decision boundaries $(-3/2, \sqrt{3^2-1}/3)$ and $(3/2, \sqrt{3^2-1}/3)$ respectively such that classifying with
$x_2>0$ is accurate, but the optimal linear classifier on one group achieves 50\% accuracy on the other.

\begin{figure}[ht!]
      \vspace{-5pt}
  \centering
  \includegraphics[scale=0.65]{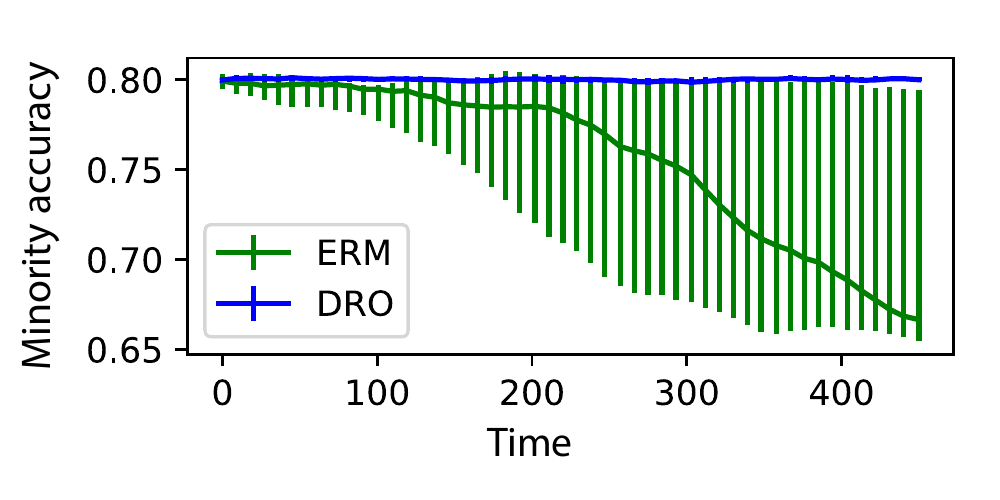}
  \caption{Disparity amplification in Figure \ref{fig:exone} is corrected by DRO. Error bars indicate quartiles over 10 replicates.}
      \vspace{-5pt}
  \label{fig:exonedro}
\end{figure}

\begin{figure}[h!]
  \centering
        \vspace{-5pt}  
  \includegraphics[scale=0.62]{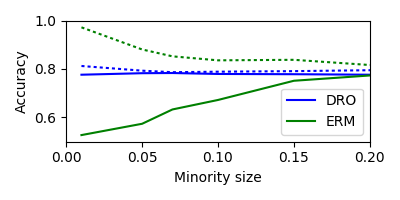}
  \caption{Classifier accuracy as a function of group imbalance. Dotted lines show accuracy on majority group.}
  \label{fig:classimb}
        \vspace{-10pt}
  \end{figure}

At each round we fit a logistic regression classifier using ERM or DRO and gradient descent, constraining the norm of the weight vector to 1. Our dynamics follow Definition \ref{def:dynamics} with $\nu(x) = 1-x$,  $\risk$ as the zero-one loss, and $b_k= 1000$. The DRO model is trained using the dual objective with logistic loss, and $\eta=0.95$, which was the optimal dual solution to $\alpha_{\min} = 0.2$. The results do not qualitatively change for choices of $\alpha_{\min} < 0.5$, and we show that we obtain control even for group sizes substantially smaller than 0.2 (Figure \ref{fig:classimb}).

Figure \ref{fig:exonedro} shows that ERM is unstable
and the minority group rapidly loses accuracy beyond $300$ rounds on most runs.
In contrast, DRO is stable, and maintains an accuracy of $0.8$.

This stability is due to the fact that the regularized loss for DRO prevents
small losses in the minority fraction from amplifying, as we discuss in Proposition \ref{prop:dynamics}.
Even when the minority fraction falls as low as $1\%$, the DRO loss ensures that
the accuracy of this minority fraction remains at $75\%$ accuracy
(Figure \ref{fig:classimb}). 

\subsection{Autocomplete task}

We now present a real-world, human evaluation of user retention and
satisfaction on a text autocomplete task. The task consists of the prediction
of next words in a corpus of tweets built from two estimated demographic
groups, African Americans and White Americans
\citep{blodgett2016}. There are several distinguishing linguistic patterns
between tweets from these groups, whose language dialects we henceforth refer
to as African-American English (AAE) and Standard-American English (SAE),
respectively, following the nomenclature in \citet{blodgett2016}. Our overall
experimental design is to measure the retention rate $\nu$ and risk $\risk$
for various choices of demographic proportions
$(\alpha_{\text{AAE}},\alpha_{\text{SAE}})$ and simulate the implied dynamics,
since running a fully online experiment would be prohibitively expensive.

For both ERM and DRO, we train a set of five maximum likelihood bigram language models on a corpus with 366,361 tweets total and a $f \in \{0.1, 0.4, 0.5 ,0.6, 0.9\}$ fraction of the tweets labeled as AAE. This results in 10 possible autocomplete systems a given Mechanical Turk user can be assigned to during a task.

To evaluate the retention and loss for AAE and SAE separately, a turk user is  assigned 10 tweets from either the held out AAE tweets or SAE tweets, which they must replicate using a web-based keyboard augmented by the autocomplete system. This assignment of a turk user to a demographic group simulates the situation where a user from a particular demographic group attempts to use the autocomplete system to write a tweet. Details of the autocomplete task are included in the supplement.

After completing the task, users were asked to fill out a survey which included a rank from 1 to 5 on their satisfaction with the task, and a yes/no question asking whether they would continue to use such a system. We assign 50 users to each of the two held out set types and each of the 10 autocomplete models, resulting in 1,000 users' feedback across autocomplete models and assigned demographics. 

The response to whether a user would continue to use the autocomplete system
provides samples $\nu(\risk_K(\alpha))$ with $n=366361$ and each of possible
demographic proportions $\alpha$. The user satisfaction survey provides a
surrogate for $\risk_K(\alpha)$ at these same points. We interpolate $\nu$ and
$\risk_K$ to $\alpha \in [0,1]$ via isotone regression which then allows us to
simulate the user dynamics and satisfaction over time using Definition
\ref{def:dynamics}. We estimate variability in these estimates via bootstrap
replicates on the survey responses.

Our results in Figure \ref{fig:auto1} show an improvement in both minority satisfaction and retention rate due to DRO: we improve the median user satisfaction from 3.7 to 4.0 and retention from 0.7 to 0.85, while only slightly decreasing the SAE satisfaction and retention. Implied user counts follow the same trend with larger differences between groups due to compounding.

Counterintuitively, the minority group has higher satisfaction and retention under DRO. Analysis of long-form comments from Turkers suggest this is likely due to users valuing the model's ability to complete slang more highly than completion of common words and indicates a slight mismatch between our training loss and human satisfaction with an autocomplete system. 


\section{Discussion}
\label{sec:relworks}

In this work we argued for a view of loss minimization as a distributive
justice problem and showed that ERM often results in disparity amplification
and unfairness. We demonstrate that DRO provides a upper bound on the risk
incurred by minority groups and performs well in practice. Our proposed algorithm is
straightforward to implement, and induces distributional robustness, which can
be viewed as a benefit in and of itself.

Our arguments against ERM and in favor of minority risk minimization 
mirror Rawls' arguments against utilitarianism, and thus inherit the critiques
of Rawlsian distributive justice. Examples of such critiques are the focus on
an abstract worst-off group rather than demographic groups or individuals
\cite{altham1973}, extreme risk-aversion \cite{mueller1974}, and
utilitarianism with diminishing returns as an alternative
\cite{harsanyi1975}. In this work, we do not address the debate on
the correctness of Rawlsian justice \citep{rawls2001}, and leave finding a suitable philosophical framework for loss minimization to future work.

There are two large open questions from our work.
First, as fairness is fundamentally a causal question, observational approaches such as DRO can only hope to control limited aspects of fairness. The generality with which our algorithm can be applied also limits its ability to enforce fairness as a constraint, and thus our approach here is unsuitable for high-stakes fairness applications such as classifiers for loans, criminality, or admissions. In such problems the implied minorities from DRO may differ from well-specified demographic groups who are known to suffer from historical and societal biases. This gap arises due to looseness in the DRO bound \cite{hu2018does}, and could be mitigated using smoothness assumptions \cite{dwork2012}.

Second, distributional robustness proposed here runs counter to classical robust estimation for rejecting outlier samples, as high loss groups created by an adversary can easily resemble a minority group.
Adversarial or high-noise settings loosen the DRO upper bound substantially, and it is an open question whether it is possible to design algorithms which are both fair to unknown latent groups and robust.

\noindent \textbf{Reproducibility:} Code to generate results available on the CodaLab platform at \url{https://bit.ly/2sFkDpE}.

\noindent \textbf{Acknowledgements:} This work was funded by an Open Philanthropy Project Award. 

\bibliography{patentref,bib,refdb/all}
\bibliographystyle{icml2018}

\onecolumn
\newpage

\appendix
\section{Appendix}

\newcommand{\maxev}{\rho_{\max}}

\subsection{Proof of Proposition~\ref{prop:unstable}}
\label{sec:proof-of-instability}

We prove the following more general result.
\begin{proposition}
  Let $\lambda^* = \Phi(\lambda^*)$ be a fixed point, and $\theta^* = \arg\min_\theta \E_{\sum_k \propk^* P_k}[\loss(\param; \statrv)]$ be the population minimizer.
  
  Define $H_{\risk}(\propvar^*)$ as the positive definite Hessian of the expected risk with $\propk^* \propto \lambda^*_k$.

  Further, let $\nabla L$ define the per-group parameter gradients at $\theta^*$,
  \[\nabla L = \begin{bmatrix}
      \nabla_\param \E_{P_1}[\loss(\hat{\param}; \statrv)] \\
      \vdots\\
      \nabla_\param \E_{P_k}[\loss(\hat{\param}; \statrv)]\\
    \end{bmatrix}.
  \]
  $\lambda^*$ is stable whenever the absolute value of the maximum eigenvalue $\maxev$ obeys 
  \begin{equation*}
    \maxev\bigg( \diag(\nu(\risk(\theta(\lambda^*)))) 
    - \diag(\lambda^* \nu'(\risk(\theta(\lambda^*)) 
    \nabla L H_{\risk}(\propvar^*)^{-1} \nabla L^\top \left( \frac{I}{\sum_k \lambda^*_k} - \frac{\mathbf{1} \lambda^{*\top}}{(\sum_k\lambda^*_k)^2}\right) \bigg) < 1.
  \end{equation*}
\end{proposition}

\begin{proof}
  A necessary and sufficient condition for stability of a discrete time dynamical system is that the Jacobian of the forward map $\Phi$ has eigenvalues with absolute value strictly less than 1.

  Computing the Jacobian we have:
  \begin{equation*}
    \mathbf{J}_{\Phi}(\lambda^*) = \diag(\nu(\risk_k(\lambda^*))) 
    + \diag(\lambda^* \nu'(\risk(\theta(\lambda^*))))\mathbf{J}_{\risk\circ \theta \circ \alpha^*}(\lambda^*).
  \end{equation*}
  Now we must compute the Jacobian of the risk with respect to the population fraction. To do this, we apply the chain rule and separately analyze three  Jacobians:$\risk$  with respect to $\theta$, $\theta$ with respect to $\propvar^*$, and $\propvar^*$ with respect to $\lambda^*$

  By strong convexity of $\loss$, 
  \[\mathbf{J}_{\theta}(\propvar^*) = - H_{\risk}(\propvar^*)^{-1} \nabla L^\top.\]
  Where $H_{\risk}(\propvar^*)^{-1}$ is the Hessian of the population risk and $\nabla L$ is
  \[\nabla L = \begin{bmatrix}
      \nabla_\theta \E_{P_1}[\loss(\hat{\param}; \statrv)] \\
      \vdots\\
      \nabla_\theta \E_{P_k}[\loss(\hat{\param}; \statrv)]\\
    \end{bmatrix}.
  \]
  The Jacobian of the risks with respect to change in $\theta$ is
  \[\mathbf{J}_{\risk}(\theta) = \nabla L.\]
  The Jacobian of the population fraction with respect to $n$ is
  \[\mathbf{J}_{\propvar^*}(\lambda^*) = \left(\frac{I}{\sum_k \lambda^*_k} - \frac{\mathbf{1}\lambda^{*\top}}{(\sum_k \lambda^*_k)^2}\right)\]
  By the chain rule, we obtain the overall claim
  \begin{equation*}
    \mathbf{J}_{\Phi}(\lambda^*) = \diag(\nu(\risk_k(\lambda^*))) 
    - \diag(\lambda^* \nu'(\risk(\theta)))\nabla L H_{\risk}(\propvar^*)^{-1} \nabla L^\top \left(\frac{I}{\sum_k \lambda^*_k} - \frac{\mathbf{1}\lambda^{*\top}}{(\sum_k \lambda^*_k)^2}\right)
  \end{equation*}
\end{proof}

 \subsection{Proof of Corollary~\ref{cor:symmetry}}
 \label{sec:proof-of-symmetry}

             Recall that the instability criteria is 
                  \begin{equation*}
        \maxev\bigg( \diag(\nu(\risk(\theta(\lambda^*)))) 
        - \diag(\lambda^* \nu'(\risk(\theta(\lambda^*)) 
        \nabla L H_{\risk}(\propvar^*)^{-1} \nabla L^\top \left( \frac{I}{\sum_k \lambda^*_k} - \frac{\mathbf{1} \lambda^{*\top}}{(\sum_k\lambda^*_k)^2}\right) \bigg) > 1.
      \end{equation*}
      Setting $\nu(\risk(\theta(\lambda^*))) = \nu(\risk_1)$ and $\lambda^*_k = \lambda^*_1$ we have,
      \begin{equation*}
        \maxev\bigg(- \nu'(\risk_1) \nabla L H_{\risk}(\propvar^*)^{-1} \nabla L^\top 
        \left( I/k - \mathbf{1} \mathbf{1}^{\top}/k^2\right) \bigg) > 1 - \nu(\risk_1).
      \end{equation*}
      By first order optimality conditions, and the fact that $\lambda^*_1 \hdots = \lambda^*_k$, $\nabla L^\top \mathbf{1}=0$.

      Thus, collecting terms and noting $\nu'(x) < 0$ by monotonicity of $\nu$, we have
      \[\maxev\bigg( \nabla L H_{\risk}(\propvar^*)^{-1} \nabla L^\top 
        \bigg) > \frac{1 - \nu(\risk_1)}{- \nu'(\risk_1)/k}.\]

      \subsection{Generalization of Proposition~\ref{prop:unstable} and Corollary~\ref{cor:symmetry}}
      \label{sec:generalizations}

Consider the more general dynamics defined by $h: (\lambda, \risk) \to \mathbb{R}^+$ which defines the evolution of the expected number of users. 

\begin{definition}
  Let $\Phi$ be the update for the expected population size
  \[\lambda^{(t+1)}_k \defeq \Phi(\lambda^{(t)}_k) = h(\lambda^{(t)}_k,\nu(\risk_k(\param(\lambda^{(t)})))),\]
  \[\param(\lambda^{(t)}_k) = \arg\min_\param \E_{\sum_k \propk^{(t)}P_k} [ \loss(\param; \statrv)].\]
\end{definition}

Then as long as $h$ is differentiable in both arguments, we obtain an essentially identical result to before.

\begin{proposition}
  Let $\lambda^* = \Phi(\lambda^*)$ be a fixed point, and $\theta^* = \arg\min_\theta \E_{\sum_k \propk^* P_k}[\loss(\param; \statrv)]$ be the population minimizer.
  
  Define $H_{\risk}(\propvar^*)$ as the positive definite Hessian of the expected risk with $\propk^* \propto \lambda^*_k$.

  Further, let $\nabla L$ define the per-group parameter gradients at $\theta^*$,
  \[\nabla L = \begin{bmatrix}
      \nabla_\param \E_{P_1}[\loss(\hat{\param}; \statrv)] \\
      \vdots\\
      \nabla_\param \E_{P_k}[\loss(\hat{\param}; \statrv)]\\
    \end{bmatrix}.
  \]
  $\lambda^*$ is stable whenever
  \begin{equation*}
    \maxev\left( \diag\left(  \frac{\partial}{\partial \lambda} h(\lambda^*, \risk(\param(\lambda^*)))\right)
    - \diag\left( \frac{\partial}{\partial \risk} h(\lambda^*, \risk(\param(\lambda^*)))\right)
    \nabla L H_{\risk}(\propvar^*)^{-1} \nabla L^\top \left( \frac{I}{\sum_k \lambda^*_k} - \frac{\mathbf{1} \lambda^{*\top}}{(\sum_k\lambda^*_k)^2}\right) \right) < 1.
  \end{equation*}
\end{proposition}

\begin{proof}
  A necessary and sufficient condition for stability of a discrete time dynamical system is that the Jacobian of the forward map $\Phi$ has eigenvalues with absolute value strictly less than 1.

  Computing the Jacobian via total derivatives we have
  \begin{equation*}
    \mathbf{J}_{\Phi}(\lambda^*) = \diag\left(  \frac{\partial}{\partial \lambda} h(\lambda^*, \risk(\param(\lambda^*)))\right)
    + \diag\left( \frac{\partial}{\partial \risk} h(\lambda^*, \risk(\param(\lambda^*)))\right) \mathbf{J}_{\risk\circ \theta \circ p}(\lambda^*).
  \end{equation*}
  The Jacobian term remains identical to before, which completes the proof. 
\end{proof}

The Corollary follows from this derivation:

\begin{corollary}[Counterexample under symmetry]\label{cor:symmetry}

  Let $\lambda^*_1 = \hdots \lambda^*_k$ be a fixed point with
  $\risk_1 = \hdots \risk_k$, and define $\frac{\partial h}{\partial \lambda} \big|_{\lambda=\lambda^*} = \frac{\partial}{\partial \lambda} h(\lambda^*_1, \risk_1)$ and $\frac{\partial h}{\partial \risk} \big|_{\lambda=\lambda^*} = \frac{\partial}{\partial \risk} h(\lambda^*_1, \risk_1)$. For any strongly convex loss,
\[\maxev\bigg( \nabla L H_{\risk}(\propvar^*)^{-1} \nabla L^\top 
        \bigg) > \frac{1 - \frac{\partial h}{\partial \lambda} \big|_{\lambda=\lambda^*}}{- \frac{\partial h}{\partial \risk} \big|_{\lambda=\lambda^*}}k\lambda^*_1.\]
\end{corollary}
\begin{proof}
  Recall that the instability criteria is
    \begin{equation*}
    \maxev\left( \diag\left(  \frac{\partial}{\partial \lambda} h(\lambda^*, \risk(\param(\lambda^*)))\right)
    - \diag\left( \frac{\partial}{\partial \risk} h(\lambda^*, \risk(\param(\lambda^*)))\right)
    \nabla L H_{\risk}(\propvar^*)^{-1} \nabla L^\top \left( \frac{I}{\sum_k \lambda^*_k} - \frac{\mathbf{1} \lambda^{*\top}}{(\sum_k\lambda^*_k)^2}\right) \right) < 1.
\end{equation*}
Let $\frac{\partial h}{\partial \lambda} \big|_{\lambda=\lambda^*} = \frac{\partial}{\partial \lambda} h(\lambda^*_1, \risk_1(\param(\lambda^*)))$ and $\frac{\partial h}{\partial \risk} \big|_{\lambda=\lambda^*} = \frac{\partial}{\partial \risk} h(\lambda^*_1, \risk_1(\param(\lambda^*)))$. Then following the same derivation as  earlier and using the monotonicity of $h$ in the second argument gives
\[\maxev\bigg( \nabla L H_{\risk}(\propvar^*)^{-1} \nabla L^\top 
        \bigg) > \frac{1 - \frac{\partial h}{\partial \lambda} \big|_{\lambda=\lambda^*}}{- \frac{\partial h}{\partial \risk} \big|_{\lambda=\lambda^*}}k\lambda^*_1.\]
\end{proof}

This is essentially in the same spirit as our earlier corollary, but requires further assumptions on $h$ in order to interpret. Generally we expect $\frac{\partial h}{\partial R}$ to be on the order of $\lambda$ as long as risk affects users independently, and $\frac{\partial h}{\partial \lambda}$ is upper bounded by the maximum implied retention rate.

\subsection{Proof of Proposition~\ref{prop:dro-includes-mixture}}
\label{sec:proof-of-dro-includes-mixture}

Since $P_k$ is a mixture component of $P$
($P = \propk P_k + \cdots$),
\begin{align*}
  \dchi{P_k}{P^{(t)}} &= \int_x \left(\frac{P_k^{(t)}(x)}{P^{(t)}(x)}-1 \right)^2 P^{(t)}(x) dx \\
                      &\leq \int_x \left(\frac{1}{
                        \propk^{(t)}}-1 \right)^2 P^{(t)}(x) dx \\
                      &= r_k.
\end{align*}
We just showed that $P_k \in \sB(P, r_k)$.
Since the sup is over all $Q \in \sB(P, r_k)$, the upper bound follows.

\subsection{Proof of Proposition~\ref{prop:dynamics}}
\label{sec:proof-of-dynamics}

By assumption,
\begin{align*}
  \propk^{(t+1)}  & \geq  \frac{\lambda^{(t)}\alphamin \nu(\risk_k(\param^{(t)}))+b_k}{\sum_k \lambda^{(t)}\propk^{(t)} \nu(\risk_k(\param^{(t)}))+b_k}.
\end{align*}
We will show
\[\frac{\lambda^{(t)}\alphamin \nu(\risk_k(\param^{(t)}))+b_k}{\sum_k \lambda^{(t)}\propk^{(t)} \nu(\risk_k(\param^{(t)}))+b_k} > \alphamin,\]
which is equivalent to
\[ \nu(\risk_k(\param^{(t)})) \geq \sum_k \propk^{(t)} \nu(\risk_k(\param^{(t)}))+ \frac{ \sum_k b_k - b_k/\alphamin}{\lambda^{(t)}}.\]

By the assumption that $\nu(\risk_k(\theta^{(t)})) < \nu_{\max}$,
\[\sum_k \propk^{(t)} \nu(\risk_k(\param^{(t)}))+ \frac{ \sum_k b_k - b_k/\alphamin}{\lambda^{(t)}} \leq
  \nu_{\max} + \frac{\sum_k b_k}{\lambda^{(t)}} \left( 1- \frac{b_k}{\alphamin \sum_k b_k} \right).\]

Using $\frac{b_k}{\sum_k b_k} \geq \alphamin$ and $\lambda^{(t)} \leq \frac{\sum b_k}{1-\nu_{\max}}$ the above simplifies to
\[\sum_k \propk^{(t)} \nu(\risk_k(\param^{(t)}))+ \frac{ \sum_k b_k - b_k/\alphamin}{\lambda^{(t)}} \leq \nu_{\max} + (1-\nu_{\max})\left(1-\frac{b_k}{\alphamin \sum_k b_k}\right).\]
Thus, a sufficient condition for our proposition is 
 \[ \risk_k(\param^{(t)}) \leq \nu^{-1}\left(1 - \frac{(1-\nu_{max})b_k}{\alphamin \sum_k b_k}\right).\].

\section{Amazon Mecahnical Turk task description}

The Amazon Mechanical Turk experiment modeling user retention in an autocomplete system is detailed below. The experiment design consists of a total of 1000 HITs ("Human Intelligence Tasks" on Mechanical Turk) consisting  of 2 user replicates $\times$ 5 values of $\alpha$ $\times$ 2 models (DRO/ERM) $\times$ 25 sets of 10 tweets from the test set $\times$ two test sets (AAE/SAE).

\begin{figure}[h!]
  \centering
  \includegraphics[page=1,scale=0.5]{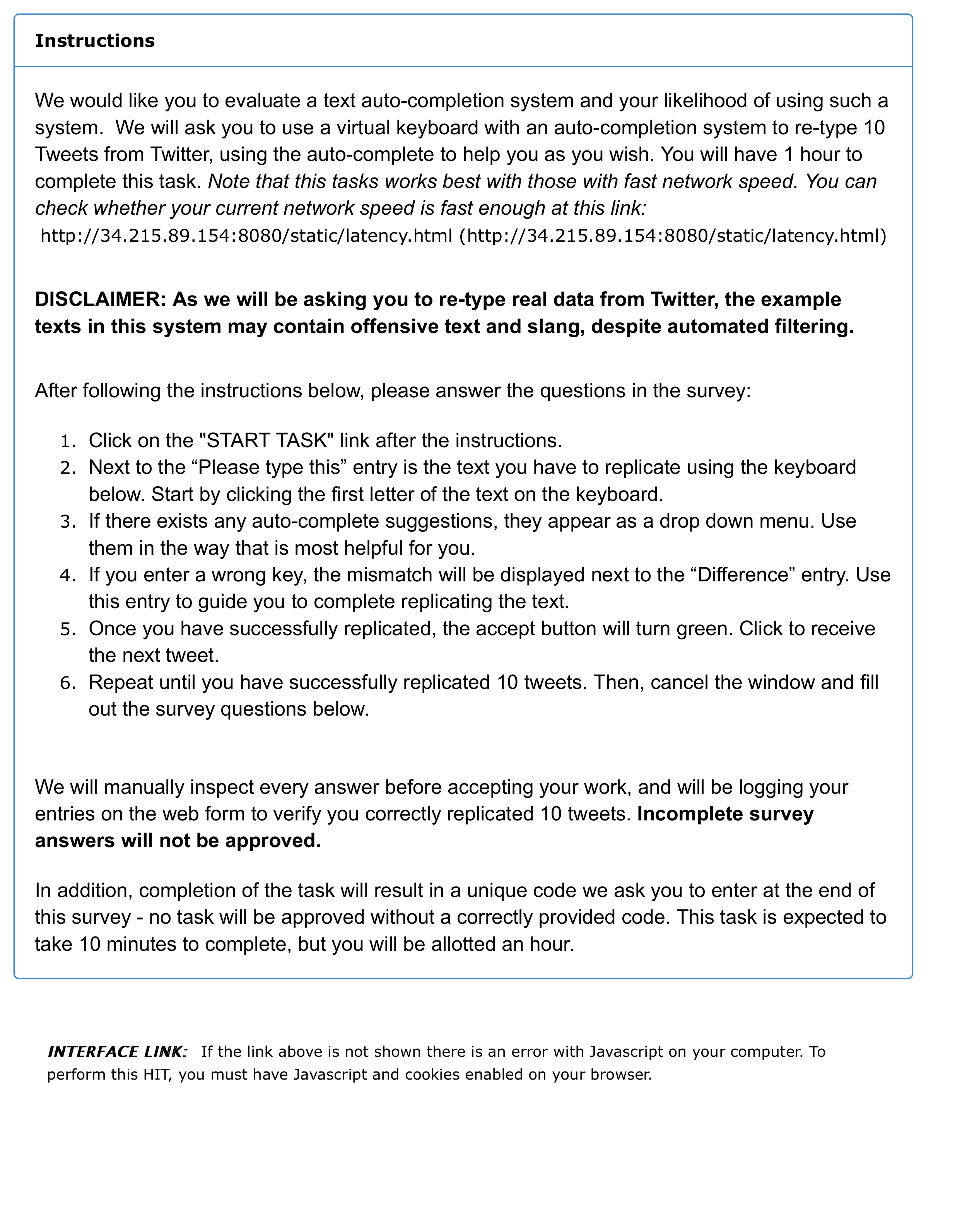}
  \caption{Task instructions on mechanical turk}
  \label{fig:taskdesc}
  \end{figure}

\begin{figure}[h!]
  \centering
  \includegraphics[scale=0.5]{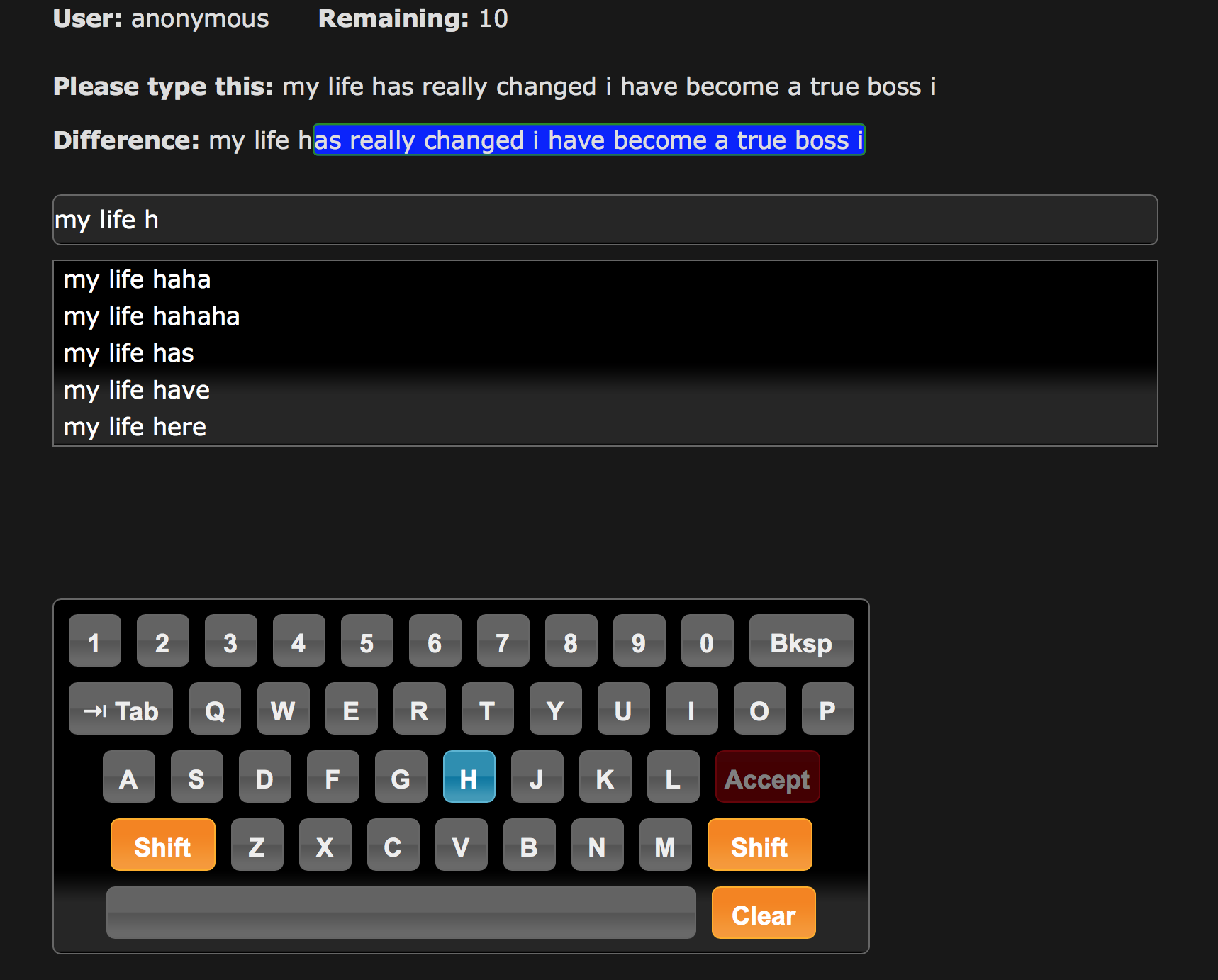}
  \caption{Autocomplete task interface on Amazon mechanical turk.}
  \label{fig:autocomp}
  \end{figure}

\begin{figure}[h!]
  \centering
  \includegraphics[page=2,scale=0.5]{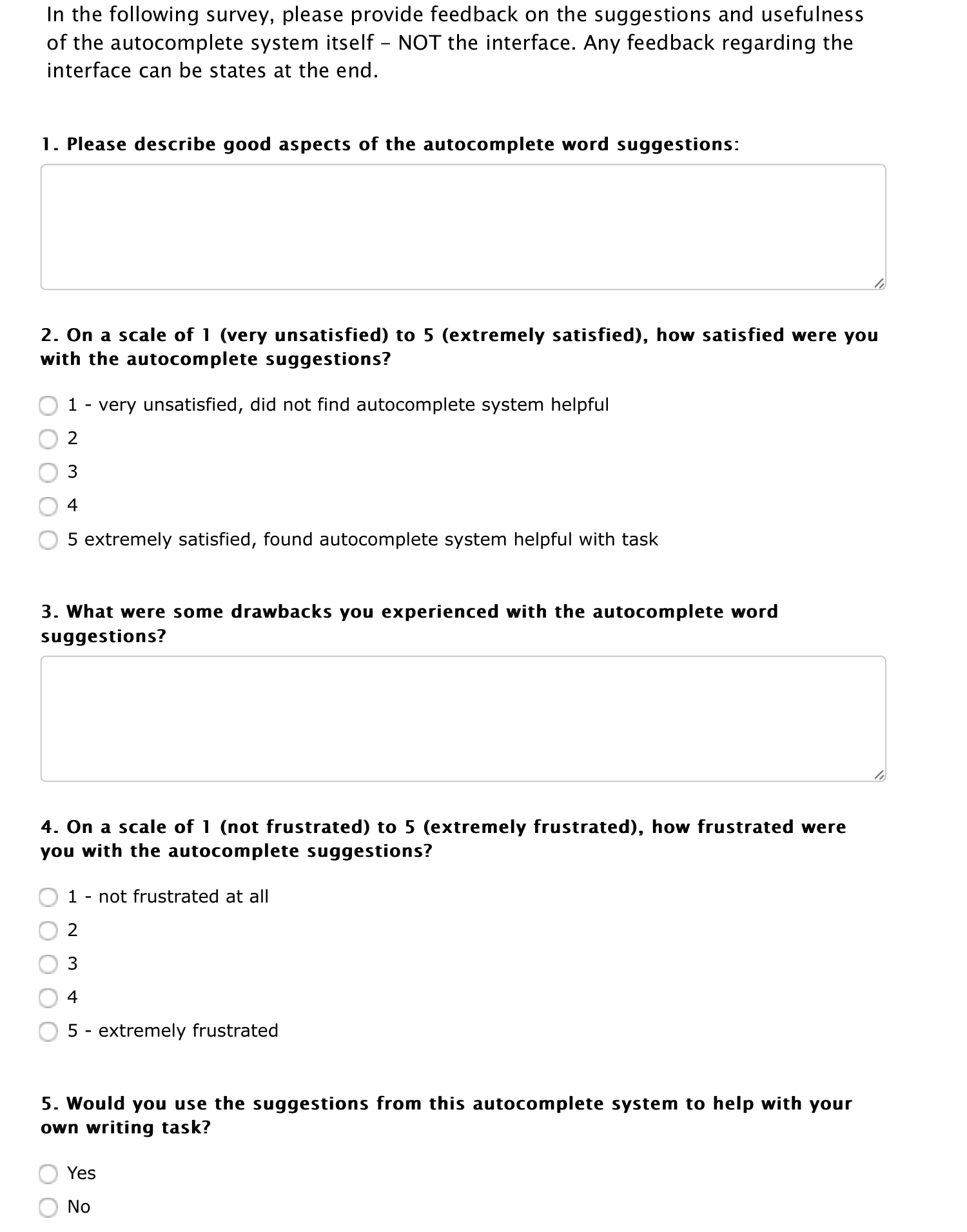}
  \caption{Survey, page 1, Q1-Q4 are quality control verification questions. Q5 measures retention}
  \label{fig:surv1}
  \end{figure}

\begin{figure}[h!]
  \centering
  \includegraphics[page=3,scale=0.5]{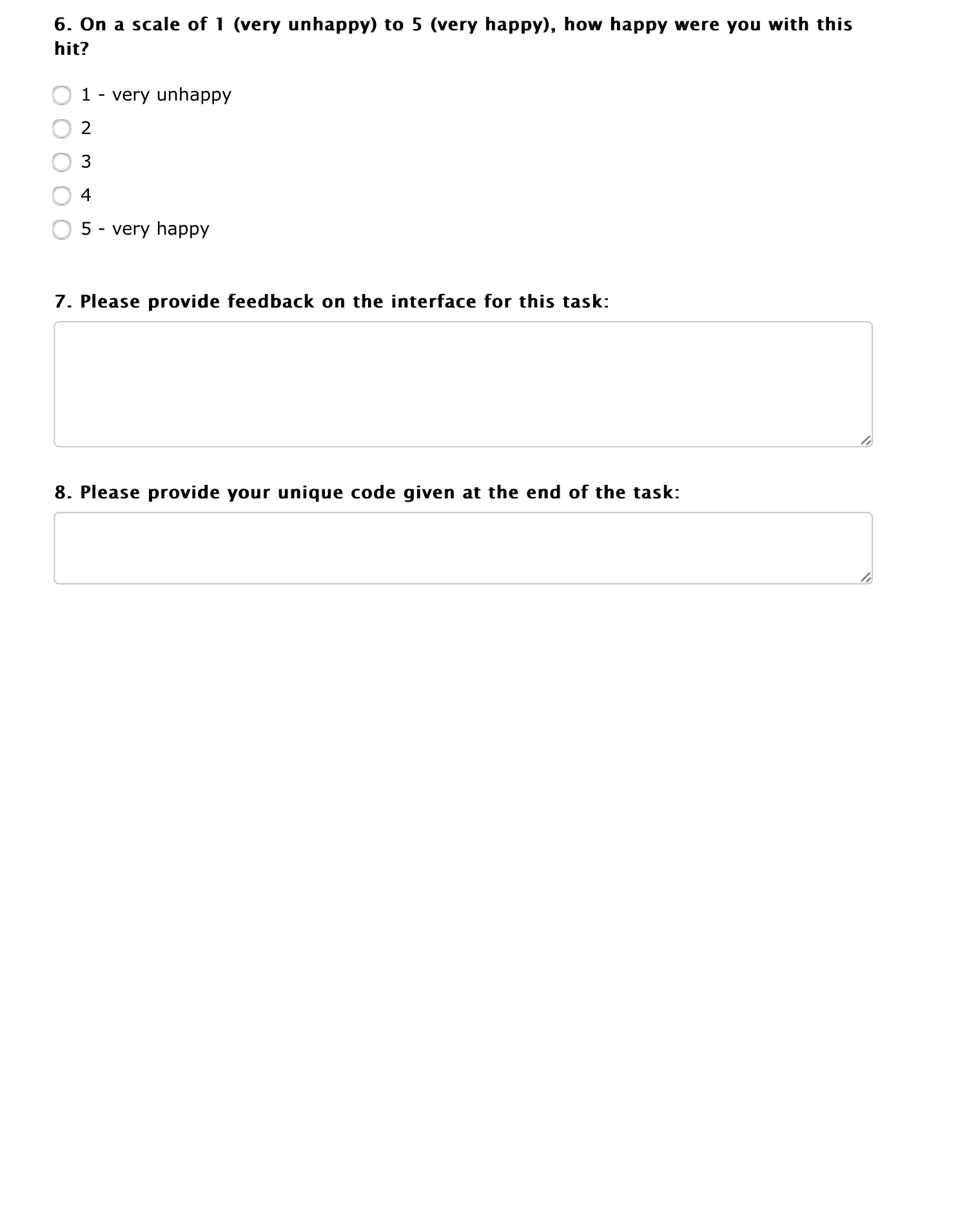}
  \caption{Survey, page 2, Q2 measures satisfaction, Q7 is used to measure issues with the HIT, and Q8 is used to ensure users completed the autocomplete task}
  \label{fig:surv2}
  \end{figure}

For each HIT the task, users are provided the description given in Figure \ref{fig:taskdesc} . Users are then taken to a separate autocomplete website, where they are asked to replicate 10 tweets using a software keyboard shown in Figure \ref{fig:autocomp}. In this interface, users must use the mouse and the software keyboard to type the target sentence, while also being given an autocomplete system for next word prediction based on the two models. The autocomplete  system appears through a dropdown as users begin typing. After completion to the task, users are prompted to fill out a survey in Figures \ref{fig:surv1},\ref{fig:surv2}. The first four questions are quality control questions designed to identify Turkers who were low effort (empty entries in Q1/Q3) or inconsistent (Q2 inconsistent with Q4). Moreover, low-quality HITS could easily be identified due to a user's refusal to select either yes or no to Q5. We used this as our metric for filtering which users would be considered in the analysis. Q6 is our overall satisfaction metric shown in the main paper.

\end{document}